# Gray-box Adversarial Training

Vivek B.S., Konda Reddy Mopuri, and R. Venkatesh Babu

Indian Institute of Science, Bangalore, India
{svivek,kondamopuri,venky}@iisc.ac.in

**Abstract.** Adversarial samples are perturbed inputs crafted to mislead the machine learning systems. A training mechanism, called adversarial training, which presents adversarial samples along with clean samples has been introduced to learn robust models. In order to scale adversarial training for large datasets, these perturbations can only be crafted using fast and simple methods (e.g., gradient ascent). However, it is shown that adversarial training converges to a degenerate minimum, where the model appears to be robust by generating weaker adversaries. As a result, the models are vulnerable to simple black-box attacks.

In this paper we, (i) demonstrate the shortcomings of existing evaluation policy, (ii) introduce novel variants of white-box and black-box attacks, dubbed "gray-box adversarial attacks" based on which we propose novel evaluation method to assess the robustness of the learned models, and (iii) propose a novel variant of adversarial training, named "Gray-box Adversarial Training" that uses intermediate versions of the models to seed the adversaries. Experimental evaluation demonstrates that the models trained using our method exhibit better robustness compared to both undefended and adversarially trained models.

**Keywords:** adversarial perturbations, attacks on machine learning models, adversarial training, robust machine learning models

## 1 Introduction

Machine learning models are observed ([1, 2, 4, 7, 17, 20]) to be susceptible to *adversarial examples*: samples perturbed with mild but structured noise to manipulate model's output. Further, Szegedy *et al.* [20] demonstrated that adversarial samples are transferable across multiple models, i.e., samples crafted to mislead one model often fool other models also. This will enable to launch simple black-box attacks [12, 18] on the models deployed in real world. These methods to generate adversarial samples, generally known as *adversaries*, range from simple gradient ascent [4] to complex optimization procedures (e.g., [14]).

Augmenting the training data with adversarial samples, known as *Adversarial Training (AT)* [4, 20] has been introduced as a simple defense mechanism against these attacks. In the adversarial training regime, models are trained with mini-batches comprising of both clean and adversarial samples, typically obtained from the same model. It is shown by Madry *et al.* [13] that adversarial training helps to learn models robust to white-box attacks, provided the perturbations computed during the training closely maximize the model's loss. However, in order to scale adversarial training for large datasets, the perturbations can only be crafted with fast and simple methods such as single-step FGSM [4, 9], an attack based on linearization of the model's loss. Tramèr *et al.* [21] demonstrated that adversarial training with single-step attacks leads to a degenerate minimum where linear approximation of model's loss is not reliable. They revealed that the model's decision surface exhibits sharp curvature near the data points which leads to overfitting in adversarially trained models. Thus, (i) adversarially trained models using single-step attacks remain susceptible to simple attacks, and (ii) perturbations crafted on undefended models transfer and form black-box attacks.

Tramèr *et al.* [21] proposed to decouple the adversary generation process from the model parameters and to increase the diversity of the perturbations shown to the model during training. Their training mechanism, called *Ensemble Adversarial Training (EAT)*, incorporates perturbations from multiple (e.g., $N$ different) pre-trained models. They showed that $EAT$ enables to learn models with increased robustness against black-box attacks.

However, $EAT$ has severe drawbacks in presenting diverse perturbations during the training. Since they augment the white-box perturbations (from the model being learned) with black-box perturbations from an ensemble of different pre-trained models, it is required to train those models before we start learning a robust model. Therefore, the computational cost increases linearly with the population of the ensemble. Because of this, the experiments presented in [21] have a maximum of 4 members in the ensemble. Though it is argued that diverse set of perturbations is important to learn robust models, $EAT$ fails to efficiently bring diversity to the table.

Unlike $EAT$, we demonstrate that it is feasible to efficiently generate diverse set of perturbations and augment the white-box perturbations. Further, utilizing these additional perturbations we learn models that are significantly robust compared to those learned with vanilla and ensemble adversarial training ($EAT$). The major contributions of this work can be listed as follows:

– We bring out an important observation that the pseudo robustness of an adversarially trained model is due to the limitations in the existing evaluation procedure.
– We introduce a novel evaluation procedure via robustness plots (3.2) and a derived metric "Worst-case Performance ($A_w$)" that can assess the susceptibility of the learned models. For that, we present variants of the white-box and black-box attacks, termed "Gray-box adversarial attacks" that can be launched by temporally evolving intermediate models. Given the efficiency to generate and the ability to examine the robustness, we strongly recommend the community to consider robustness plots and "Worst-case Performance" as standard bench-marking for evaluating the models.
– Harnessing the above observations, we propose a novel variant of adversarial training, termed "Gray-box Adversarial Training" that uses our gray-box perturbations in order to learn robust models.

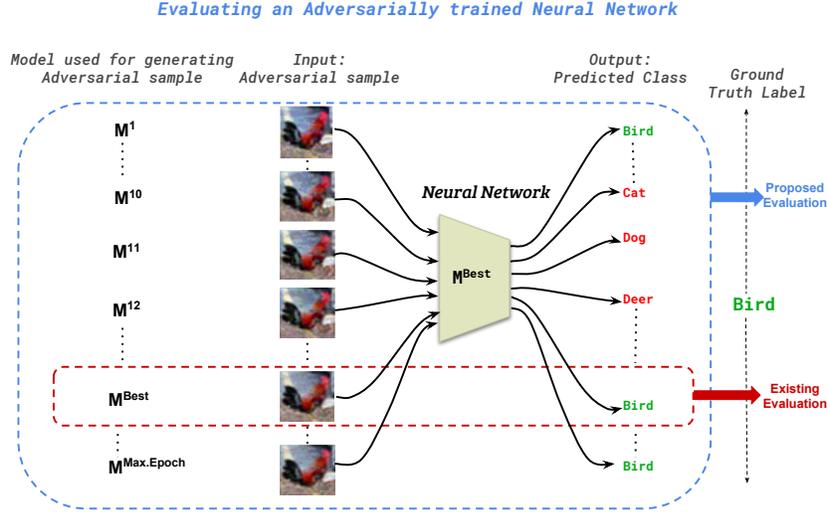

Fig. 1: Overview of existing and proposed evaluation methods for testing the robustness of *adversarially trained network* against adversarial attacks. For existing evaluation, *best model's* robustness against adversarial attack is tested by obtaining it's prediction on adversarial samples generated by itself. Whereas, for the proposed evaluation method adversarial samples are not only generated by *best model* but also by the intermediate models obtained while training.

The paper is organized as follows: section 2 introduces the notation followed in the subsequent sections of the paper, section 3 presents the drawbacks in the current robustness evaluation methods for deep networks and proposes improved procedure, section 4 hosts the experiments and results, section 5 discusses existing works that are relevant, and section 6 concludes the paper.

## 2 Notations and terminology

In this section we define the notations followed throughout this paper:

- $x$ : clean image from the dataset.
- $x^*$ : a potential adversarial image corresponding to the image $x$.
- $y_{true}$ : ground truth label corresponding to the image $x$.
- $y_{pred}$ : prediction of the neural network for the input image $x$.
- $\epsilon$ : defines the strength of perturbation added to the clean image.
- $\theta$ : parameters of the neural network.
- $J$ : loss function used to train the neural network.
- $\nabla_x J$ : gradient of the loss $J$ with respect to image $x$.
- *Best Model* ($M^{Best}$): Model corresponding to least validation loss, typically obtained at the end of the training.
- $M_t^i$ : represents model at $i^{th}$ epoch of training, obtained when network $M$ is trained using method 't'.

## 3    Gray-box Adversarial Attacks

### 3.1    Limitations of existing evaluation method

Existing ways of evaluating an adversarially trained network consists of evaluating the best model's accuracy on adversarial samples generated by itself. This way of evaluating the networks gives false inference about their robustness against adversarial attacks. This assumes (though explicitly not mentioned) that robustness to the adversarial samples generated by the best model extends to the adversarial samples generated by the intermediate models evolved during the training, crafted via linear approximation of the loss (e.g., FGSM [4]). Clearly, this is not true, as shown by our robustness plots (see Figure 2). We show this by obtaining robustness plot which captures accuracy of best-model not only on adversarial samples generated by itself but also on adversarial samples generated by the intermediate models, which are obtained during training. Based on the above facts we propose two new ways of evaluating adversarially trained network, shown in Table 1.

### 3.2    Robustness plot and Worst-case performance

We propose a new way of evaluating the robustness of a network, which is a plot of recognition accuracy of the best model $M^{Best}$ on adversarial samples of different perturbation strengths $\epsilon$, generated by multiple intermediate models that are obtained during training. That is, performance of the model under investigation is evaluated against the adversaries of different perturbation strengths $\epsilon$, generated by checkpoints or models saved during training. Based on the source of these *saved models* which seed adversarial samples, we differentiate robustness plot into two categories.

- If the saved models and the best model are obtained from the same network and also they both have the same training procedure, then we name such *robustness plot* as *Extended White-box robustness plot*. Further, we name the attacks as "*Extended White-box adversarial attacks*".
- Else, if the network trained or the training procedure used are different, then we call such robustness plot as *Extended Black-box robustness plot* and such attacks as "*Extended Black-box adversarial attacks*".

In general, we call these attacks as *Gray-box adversarial attacks*. We believe it is intuitive to call the proposed attacks as "Extensions" to existing white and black box attacks. White-box attack means the attacker has full access to the target model: architecture, parameters, training data and procedure. Typically, source model that creates adversaries is same as the target model under attack. Whereas, "extended white-box" attack means, some aspects of the setup are known while some are not. Specifically, the architecture and the training procedure of the source and target model are same while the model parameters differ. Similarly, black-box attack means the scenario where the attacker has no information about the target such as architecture, parameters, training procedure, etc. Generally, the source model would be a fully trained model that has different architecture (and hence parameters) compared to the target model. However, the "extended black-box" attack mean, the source model can be a partially trained model having different network architecture. Note that it is not very different compared to the existing black-box attack except the source model can now be a partially trained model. We jointly call these two extended attacks as "Gray-box Adversarial Attacks". Table 1 lists the definitions of both the existing and our extended attacks with the notation we introduced earlier in the paper.

Table 1: List of the adversarial attacks. Note that the subscript denotes the training procedure how the model is trained, superscript denotes the training epoch at which the model is considered. $M^i$ denotes intermediate model and $M^{Best}$ denotes the best model.

| Source | Target | Name of the attack |
|---|---|---|
| $M_{Adv.}^{Best}$ | $M_{Adv.}^{Best}$ | White-box attack |
| $X_{normal}^{Best}$ | $M_{Adv.}^{Best}$ | Black-box attack |
| $M_{Adv.}^{i}$ $i = 1, \ldots,$ MaxEpoch | $M_{Adv.}^{Best}$ | Extended White-box attack |
| $X_{Normal/Adv.}^{i}$ $i = 1, \ldots,$ MaxEpoch | $M_{Adv.}^{Best}$ | Extended Black-box attack |

**Worst-case performance** ($A_w$): We introduce a metric derived from the proposed robustness plot to infer the susceptibility of the trained model quantitatively in terms of its weakest performance. We name it "Worst-case Performance ($A_w$)" of the model, which is the least recognition accuracy achieved for a given attack strength ($\epsilon$). Ideally, for a robust model the value of $A_w$ should be high (close to its performance on clean samples).

### 3.3   Which attack to use for adversarial training, FGSM, FGSM-LL or FGSM-Rand?

Kurakin *et al.* [9] suggested to use FGSM-LL or FGSM-Rand variants for adversarial training, in order to reduce the effect of *label leaking* [9]. *Label leaking* effect is observed in adversarially trained networks, where the accuracy of the model on the adversarial samples of higher perturbation is greater than that on the adversarial samples of lower perturbation. It is necessary for adversarial training to include stronger attacks in the training process in order to make the model robust. We empirically show (in sec. 4.2 and Fig. 3) that FGSM is stronger attack compared to FGSM-LL and FGSM-Rand through robustness plots with the three different attacks for a normally trained network.

In addition, for models (of same architecture) adversarially trained using FGSM, FGSM-LL and FGSM-Rand attacks respectively, FGSM attack causes more damage compared to other two attacks. We show (in sec. 4.2 and Fig. 4) this by obtaining robustness plots with FGSM, FGSM-LL, and FGSM-Rand attacks respectively, for all three variants of adversarial training methods. Based on these observations we use FGSM for all our experiments.

**Algorithm 1:** Gray-box adversarial training of network $N$

**Input:**
    $m =$ Size of the training minibatch
    $k =$ No. of adversarial images in minibatch generated using current state of the network $N$
    $p =$ No. of adversarial images in minibatch generated using $i^{th}$ state of the network $N$
    $MaxItertion =$ Maximum training iterations
    Number of clean samples in minibatch $= m - k - p$
    Hyper-parameters EL, D and T

**1 Initialization**
  Randomly initialize network **N**
  */\*Set containing the iterations at which seed models are saved\*/*
  $AdvBag = \{\}$
  $AdvPtr=0$ */\*Pointer to elements in set 'AdvBag'\*/*
  $i=0$ */\*Refers to initial state of the network\*/*
  Initialize $LossSetPoint$ with initial training loss
  $iteration = 0$

**2 while** $iteration \neq MaxItertion$ **do**
**3**    Read minibatch $B = \{x^1, ..., x^m\}$ from training set
**4**    Generate '$k$' adversarial examples $\{x_{adv}^1, ..., x_{adv}^k\}$ from corresponding clean samples $\{x^1, ..., x^k\}$ using current state of the network $N$
**5**    Generate '$p$' adversarial examples $\{x_{adv}^{k+1}, ..., x_{adv}^{k+p}\}$ from corresponding clean samples $\{x^{k+1}, ..., x^{k+p}\}$ using $i^{th}$ state of the network $N$
**6**    Make new minibatch $B^* = \{x_{adv}^1, ..., x_{adv}^k, x_{adv}^{k+1}, ..., x_{adv}^{k+p}, x^{k+p+1}, ..., x^m\}$
**7**    */\*forward pass, compute loss, backward pass, and update parameters\*/*
**8**    Do one training step of Network $N$ using minibatch $B^*$
**9**    */\*moving average loss computed over 10 iterations\*/*
     $LossCurrentValue = MovingAverage(loss)$
**10**   */\*Logic for saving seed model \*/*
**11**   **if** $(LossSetPoint - LossCurrentValue) \geq D$ and $LossSetPoint \geq EL$ **then**
**12**      $AdvBag.add(iteration)$
**13**      SaveModel($N^{iteration}$)
**14**      $LossSetPoint = LossCurrentValue$
**15**   **end**
**16**   */\*Logic for picking saved seed model\*/*
**17**   **if** $(iteration \% T) == 0$ and $len(AdvBank) \neq 0$ **then**
**18**      $i = AdvBank[AdvPtr]$
**19**      $AdvPtr = (AdvPtr + 1) \% len(AdvBank)$
**20**   **end**
**21**   $iteration = iteration + 1$
**22 end**

### 3.4  Gray-box Adversarial Training

Based on the observations presented in sec. 3.1, we propose *Gray-box Adversarial Training* in Algorithm 1 to alleviate the drawbacks of existing adversarial train-ing. During training, for every iteration, we replace a portion of clean samples in the mini-batch with its corresponding adversarial samples which are generated not only by the current state of the network but also by one of the saved intermediate models. We use "drop in training loss" as a criterion for saving these intermediate models during training i.e., for every $D$ drop in the training loss we save the model and this process of saving the models continues until training loss reaches minimum prefixed value $EL$ (End Loss).

The intuition behind using "drop in training loss" as a criterion for saving the intermediate models is that, models substantially apart in the training (evolution) process can source different set of adversaries. Having variety of adversarial samples to participate in the adversarial training procedure makes the model robust [21]. As training progresses, network representations evolve and loss decreases from the initial value. Thus, we use the "drop in training loss" as a useful index to pick source models that can potentially generate different adversaries. We represent this quantity as $D$ in the proposed algorithm.

Ideally we would like to have an ensemble of as many different source models as possible. However, bigger ensembles would pose additional challenges such as computational, memory overheads and slow down the training process. Therefore, $D$ has to be picked depending on the trade-off between performance and overhead. Note that too small value of $D$ will include highly correlated models in the ensemble. Also, towards later stages of the training, model evolves very slowly and representations would not change significantly. So, after some time into the training, we stop augmenting the ensemble of source models. For this we define a parameter denoted as $EL$ (End Loss) which is a threshold on the loss that defines when to stop saving the intermediate models. During the training process, once the loss falls below $EL$, we stop saving of intermediate models and prevent augmenting the ensemble with redundant models.

Further, in the best case we would like to pick different saved model for every iteration during our Gray-box Adversarial Training. However, this creates bottleneck because loading a saved model at each iteration is time consuming. In order to reduce this additional overhead, we pick a saved model and use this for $T$ consecutive iterations after which we pick another saved model in a round-robin fashion. In total, we have three hyper-parameters namely, $D$, $EL$, and $T$, and we show the effect of these hyper-parameters in sec. 4.3.

## 4 Experiments

In our experiments we show results on CIFAR-100, CIFAR-10 [8], and MNIST [10] dataset. We work with WideResNet-28-10 [22] for CIFAR-100, ResNet-18 [6] for CIFAR-10 and LeNet [11] for MNIST dataset, all these networks achieve near state of the art performance on the respective dataset. These networks are trained for 100 epochs (25 epochs for LeNet) using SGD with momentum, and models are saved at the end of each epoch. For learning rate scheduling, step-policy is used. We pre-process images to be in $[0, 1]$ range, and random crop and horizontal flip are performed for data-augmentation (except for MNIST). Experiments and results on MNIST dataset are shown in supplementary document.

### 4.1  Limitations of existing evaluation method

In this subsection, we present the relevant experiments to understand the issues present in the existing evaluation method as discussed in Section. 3.1. We adversarially train, WideResNet-28-10 and ResNet-18 on CIFAR-100 and CIFAR-10 datasets respectively and while training, FGSM is used for adversarial sample generation process. After training, we obtain their corresponding Extended White-box robustness plot using FGSM attack. Figure 2 shows the obtained Ex-tended White-box robustness plot. It can be observed that adversarially trained networks are not robust to the adversaries generated by the intermediate models, which the existing way of evaluation fails to capture. It also infers that the implicit assumption of best model's robustness to adversaries generated by the intermediate models is false. We also reiterate the fact that existing adversarial training formulation does not make the network robust but makes them to generate weaker adversaries.

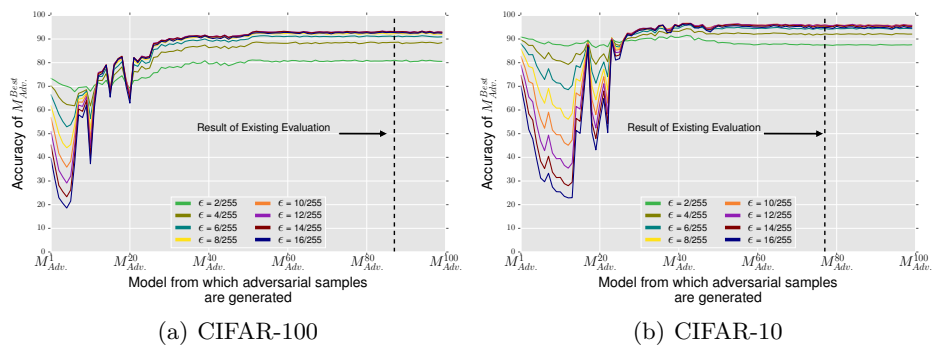

(a) CIFAR-100          (b) CIFAR-10

Fig. 2: Extended White-box robustness plots with FGSM attack, obtained for (a) WideResNet-28-10 adversarially trained on CIFAR-100 dataset, (b) ResNet-18 adversarially trained on CIFAR-10 dataset. Note that the classification accuracy of the best model ($M_{Adv}^{Best}$) is poor for the attacks generated by early models (towards origin) as opposed to that by the later models.

### 4.2  Which attack to use for adversarial training, FGSM, FGSM-LL or FGSM-Rand?

We perform normal training of WideResNet-28-10 on CIFAR-100 dataset and obtain its corresponding Extended White-box robustness plots using FGSM,

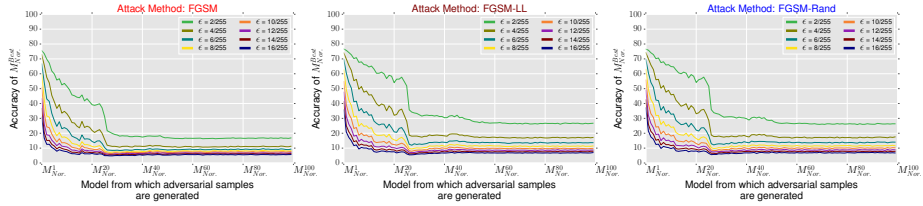

**Fig. 3:** Extended White-box robustness plots of WideResNet-28-10 normally trained on CIFAR-100 dataset, obtained using FGSM (column-1), FGSM-LL (column-2) and FGSM-Rand (column-3) attacks. Note that for a wide range of perturbations, FGSM attack causes more dip in the accuracy

FGSM-LL, and FGSM-Rand attacks respectively. Figure 3 shows the obtained plots. It is clear that FGSM attack (column-1) produces stronger attacks compared to FGSM-LL (column-2) and FGSM-Rand (column-3) attacks. That is, drop in the model's accuracy is more for FGSM attack.

Additionally, we adversarially train the above network using FGSM, FGSM-LL and FGSM-Rand respectively for adversarial sample generation process. After training, we obtain robustness plots with FGSM, FGSM-LL and FGSM-Rand attacks respectively. Figure 4 shows the obtained Extended White-box robustness plots for all the three versions of adversarial training. It is observed that the model is more susceptible to FGSM attack (column-1) compared to other two attacks. Similar trends are observed for networks trained on CIFAR-10 and MNIST datasets and corresponding results are shown in supplementary document.

### 4.3 Gray-box adversarial training

We train WideResNet-28-10 and Resent-18 on CIFAR-100 and CIFAR-10 datasets respectively, using the proposed Gray-box Adversarial Training (GAT) algorithm. In order to create strong adversaries, we chose FGSM to generate adversarial samples. We use the same set of hyper-parameters $D= 0.2$, $EL= 0.5$, and $T= (1/4^{th}$ of an epoch), for all the networks trained. Specifically, we save intermediate models for every 0.2 drop ($D$) in the training loss till the loss falls below 0.5 ($EL$). Also, each of the saved models are sampled from the ensemble, generates adversarial samples for $1/4^{th}$ of an epoch. After training, we obtain Extended White-box robustness plots and Extended Black-box robustness plots for networks trained with the Gray-box adversarial training (Algorithm 1) and for adversarially trained networks. Figure 5 shows the obtained robustness plots, it is observed from the plots that in the proposed gray-box adversarial training (row-2) there are no deep valleys in the robustness plots, whereas for the networks trained using existing adversarially training method (row-1) exhibits deep valley in the robustness plots. Table 2 presents the *worst-case accuracy* $A_w$ of the models trained using different training methods. Note that $A_w$ is significantly better for the proposed $GAT$ compared to the model trained with $AT$ and $EAT$ for a wide range of attack strength ($\epsilon$).

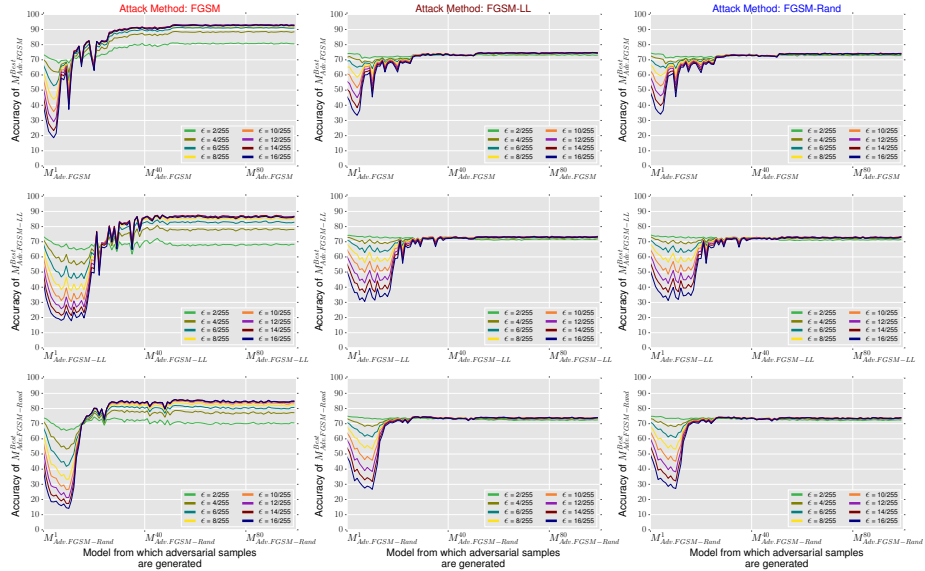

Fig. 4: Extended White-box robustness plots of WideResNet-28-10 trained on CIFAR-100 dataset using different adversarial training methods, obtained using FGSM (column-1), FGSM-LL (column-2) and FGSM-Rand (column-3) attacks. Rows represents training method used, Row-1 : Adversarial training using FGSM, Row-2 : Adversarial training using FGSM-LL and Row-3 : Adversarial training using FGSM-Rand.

**Effect of hyper-parameters**: In order to study the effect of hyper parameters, we train ResNet-18 on CIFAR-10 dataset using Gray-box adversarial training for different hyper-parameter settings. Extended White-box robustness plots are obtained for each setting with two of them fixed and the other being varied. The hyper-parameter $D$ defines the value of "drop in training loss" for which intermediate models are saved to generate adversarial samples. Figure 6(a) shows the effect of varying $D$ from 0.2 to 0.4. It is observed that for higher values of $D$, the depth and the width of the valley increases. This is because choosing higher values of $D$ might miss saving of models that are potential sources for generating stronger adversarial samples, and also choosing very low values of $D$ will results in saving large number of models that are redundant and may not be useful. The hyper-parameter $EL$ decides when to stop saving of intermediate seed models. Figure 6(b) shows the effect of $EL$ for fixed values of $D$ and $T$. We observe that as $EL$ increases the width of the valley increases since higher values of $EL$ prevents saving of potential models. Finally, the hyper-parameter $T$ decides the duration for which a member of ensemble is used after getting sampled from the ensemble to generate adversarial samples. Figure 6(c) shows the effect

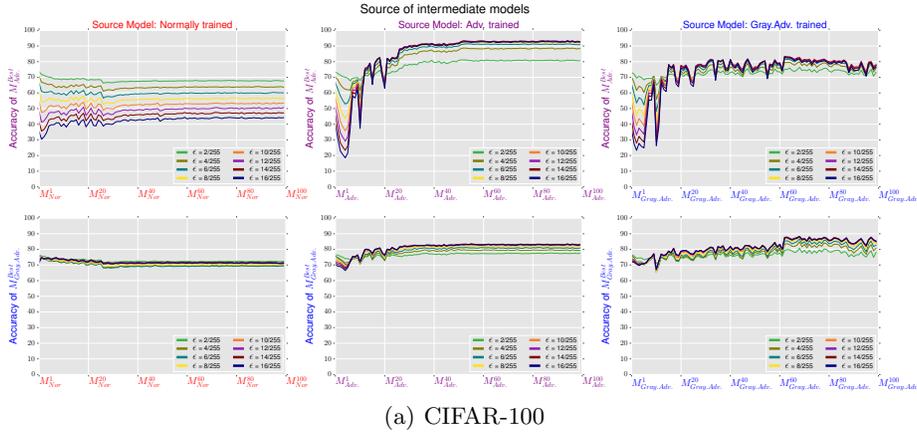

(a) CIFAR-100

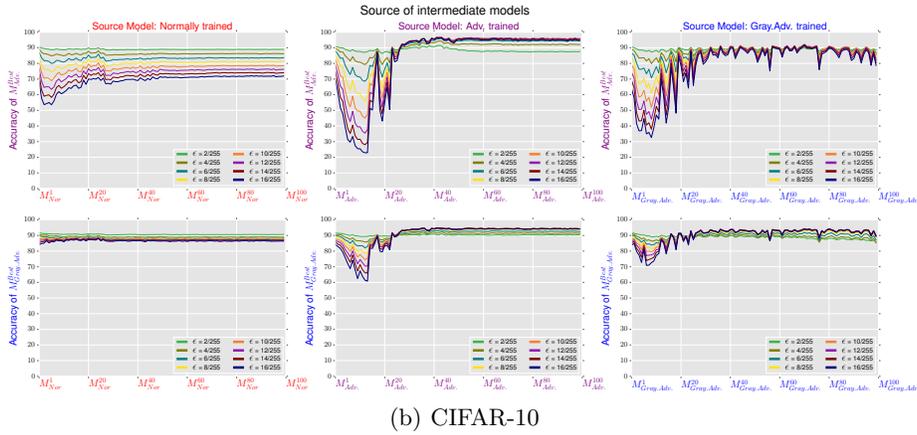

(b) CIFAR-10

Fig. 5: Robustness plots obtained using FGSM attack for (a)WideResNet-28-10 trained on CIFAR-100 and (b) ResNet-18 trained on CIFAR-10. Rows represents the training method, Row-1:Model trained adversarially using FGSM and Row-2:Model trained using Gray-box adversarial training. Adversarial samples are generated by intermediate models of Column-1:Normal training, Column-2: Adversarial training using FGSM, Column-3:Gray-box adversarial training

of varying $T$ from $1/4^{th}$ epoch to $3/4^{th}$ epoch. Note that $T$ has minimal effect on the robustness plot within that range.

### 4.4 Ensemble adversarial training

In this subsection, we compare our Gray-box Adversarial Training against Ensemble Adversarial Training [21]. We work with the networks on CIFAR-100 and CIFAR-10 datasets. Ensemble adversarial training uses fixed set of pre-trained models for generating adversarial samples along with the current state of the

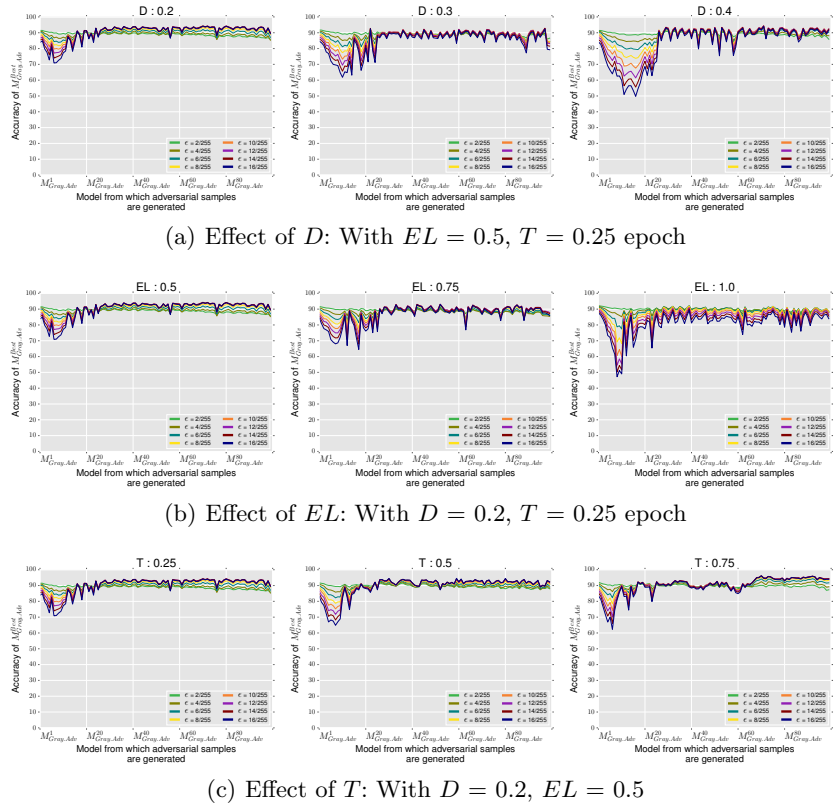

(a) Effect of $D$: With $EL = 0.5$, $T = 0.25$ epoch

(b) Effect of $EL$: With $D = 0.2$, $T = 0.25$ epoch

(c) Effect of $T$: With $D = 0.2$, $EL = 0.5$

Fig. 6: Extended White-box robustness plots of ResNet-18 trained on CIFAR-10 dataset using Gray-box adversarial training (algorithm 1), for different hyper-parameters settings. Effects of hyper-parameters are shown in (a) Effect of $D$, (b) Effect of $EL$ and (c) Effect of $T$ (measured in fraction of an epoch).

Table 2: Worst case accuracy of models trained on (a) CIFAR-100 and (b) CIFAR-10, using different training methods. For ensemble adversarial training ($EAT$) refer to section 4.4. A, B and C refers to the ensemble used in $EAT$.

(a) CIFAR-100

| Training Method | | $A_w$ for various $\epsilon$ | | | |
|---|---|---|---|---|---|
| | | 2/255 | 4/255 | 8/255 | 16/255 |
| Normal | | 20.81 | 11.58 | 7.09 | 4.8 |
| AT | | 70.04 | 62.34 | 44.04 | 18.58 |
| $EAT$ | A | 65.39 | 55.55 | 36.8 | 15.2 |
| | B | 65.45 | 54.63 | 35.65 | 14.26 |
| | C | 65.71 | 56.05 | 37.43 | 15.98 |
| $GAT$(ours) | | **70.84** | **66.5** | **65.43** | **67.41** |

(b) CIFAR-10

| Training Method | | $A_w$ for various $\epsilon$ | | | |
|---|---|---|---|---|---|
| | | 2/255 | 4/255 | 8/255 | 16/255 |
| Normal | | 38.15 | 20.84 | 12.2 | 9.34 |
| Adversarial | | 87.08 | 79.25 | 56.09 | 22.87 |
| $EAT$ | A | 83.27 | 73.4 | 53.96 | 31.01 |
| | B | 82.56 | 72.35 | 56.11 | 34.84 |
| | C | 82.64 | 73.84 | 55.19 | 32.33 |
| $GAT$(ours) | | **89.46** | **85.89** | **79.28** | **60.81** |

Table 3: Setup for ensemble adversarial training

|  | **Network to be trained** | **Pre-trained Models** | **Held-out Model** |
|---|---|---|---|
| CIFAR-100 | WideResNet-28-10(Ensemble-A) | Resnet-50, ResNet-34 | WideResNet-28-10 |
|  | WideResNet-28-10(Ensemble-B) | WideResNet-28-10, ResNet-50 | ResNet-34 |
|  | WideResNet-28-10(Ensemble-C) | WideResNet-28-10, ResNet-34 | ResNet-50 |
| CIFAR-10 | ResNet-34(Ensemble-A) | ResNet-34, ResNet-18 | VGG-16 |
|  | ResNet-34(Ensemble-B) | ResNet-34, VGG-16 | ResNet-18 |
|  | ResNet-34(Ensemble-C) | ResNet-18, VGG-16 | ResNet-34 |

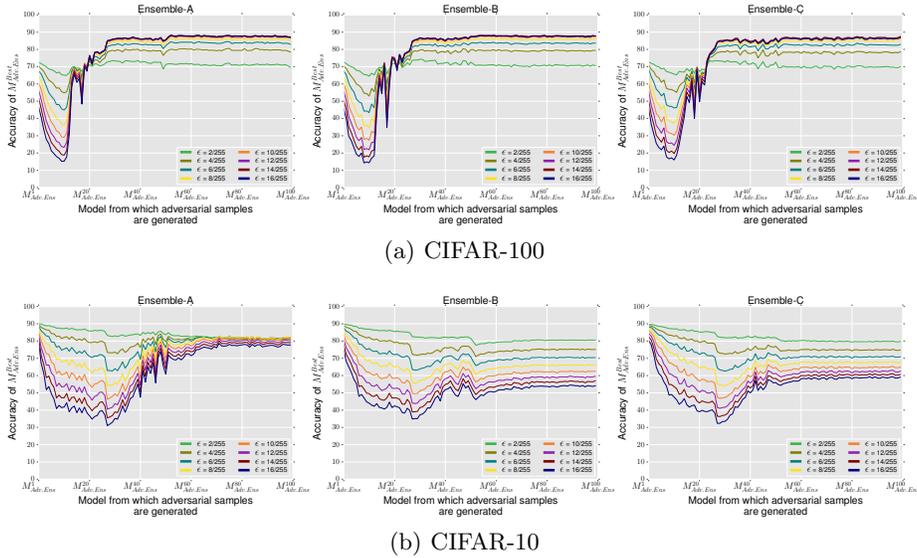

(a) CIFAR-100

(b) CIFAR-10

Fig. 7: Extended White-box robustness plots of models trained using ensemble adversarial training algorithm. (a)models trained on CIFAR-100 dataset and (b)models trained on CIFAR-10 dataset

network. For each iteration during training, the source model for generating adversaries is picked at random among the current and ensemble models. Table 3 shows the setup used for ensemble adversarial training, which contains network to be trained, pre-trained source models used for generating adversarial samples and the held-out model used for black-box attack. Figure 7 shows the Extended White-box robustness plots for the networks trained using ensemble adversarial training algorithm. Note the presence of deep and wide valleys in the plot. Whereas, the Extended White-box robustness plots for the models trained using the Gray-box adversarial training shown in figure 5 (row:2,column:3), do not have deep and wide valley. Also, because of the space restrictions, Extended Black-box robustness plots for the above trained networks using ensemble adversarial training algorithm are shown in supplementary document.

## 5   Related Works

Following the findings of Szegedy *et al.* [20], various attack methods (e.g. [3, 4, 14, 15, 16]) and various defense techniques (e.g. [4, 5, 13, 21, 19]) have been proposed. On the defense side, adversarial training shows promising results. In order to scale adversarial training [9] to large datasets, single-step attack methods (use 1st order approximation of model's loss to generate attack ) are used while training. Goodfellow *et al.* [4] observed that adversarially trained models, incurs higher loss on transferred samples than on the white-box single-step attacks. Further, Kurakin [9] observed that adversarially trained models were susceptible to adversarial samples generated using multi-step methods under white-box setting. This paradoxical behaviour is explained by Tramèr *et al.* [21] through the inaccuracy of linear approximation of the model's loss function in the vicinity of the data samples. Madry et al. [13] showed that adversarially trained model can become robust against white-box attacks, if adversaries added during training closely maximizes the model's loss. However, such methods are hard to scale to difficult tasks such as ILSVRC [9]. Tramèr *et al.* [21] showed that having diversity in the adversarial samples presented during the training can alleviate the effect of gradient masking. However, it is inefficient to have an ensemble of different pre-trained models to seed the adversarial samples. Apart from highlighting the flaws in the existing evaluation and proposing better evaluation method, we propose efficient way of generating diverse adversarial samples for learning robust models.

## 6   Conclusions

Presence of adversarial samples indicates the vulnerability of the machine learning models. Learning robust models and measuring their susceptibility against adversarial attacks are need of the day. In this paper, we demonstrated important gaps in the existing evaluation method for testing the robustness of a model against adversarial attacks. Also, we proposed a novel evaluation method called "Robustness plots" and a derived metric "Worst-case performance ($A_w$)". From the proposed evaluation method, it is observed that the existing adversarial training methods which use first order approximation of loss function for generating samples, do not make the model robust. Instead, they make the model to generate weaker adversaries. Finally, the proposed "Gray-box Adversarial Training ($GAT$)" which harnesses the presence of stronger adversaries during training, achieves better robustness against adversarial attacks compared to the existing adversarial training methods (that follow single step adversarial generation process).

GAT      15

## 7  Supplementary

Supplementary section is organized as follows:

- Section 8: Experimental results on CIFAR-10 dataset, to understand the strength of various adversarial generation methods used in adversarial training.
- Section 9: Extended Black-box robustness plots of networks trained using *Ensemble Adversarial Training* (*EAT*) on CIFAR-100 and CIFAR-10 datasets .
- Section 10: Experimental results on MNIST dataset
    - Subsection 10.1: Limitations of existing evaluation method
    - Subsection 10.2: Which attack to use for adversarial training, FGSM, FGSM-LL or FGSM-Rand ?
    - Subsection 10.3: Gray-box adversarial training ($GAT$)
    - Subsection 10.4: Ensemble adversarial training ($EAT$)

## 8  CIFAR-10: Which attack to use for adversarial training, FGSM, FGSM-LL or FGSM-Rand?

In this section we present results on models trained on CIFAR-10 dataset for experiments presented in section 4.2 of the main paper, which discusses the strength of various adversarial sample generation methods. Figure 8 shows Extended White-box robustness of normally trained ResNet-18, obtained using FGSM, FGSM-LL and FGSM-Rand attacks respectively. Figure 9 shows Extended White-box robustness plots of ResNet-18 adversarially trained using different adversarial sample generation methods. From the figures it is observed that for FGSM attack the recognition plots are closer to the bottom (x-axis) of the graph, i.e., towards the lower recognition accuracy. This means, FGSM causes more damage to the model (defended or undefended) compared to other two attacks namely FGSM-LL and FGSM-Rand attacks.

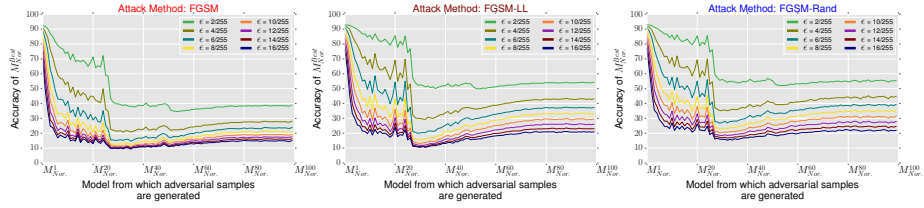

Fig. 8: Extended White-box robustness plots of ResNet-18 normally trained on CIFAR-10 dataset, obtained using attacks (Column-1) FGSM, (Column-2) FGSM-LL and (Column-3) FGSM-Rand.

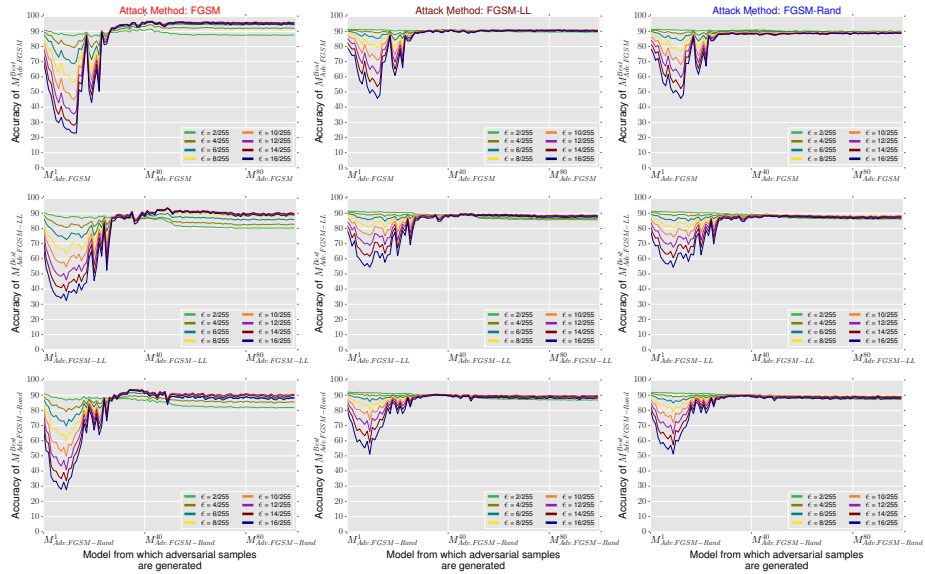

Fig. 9: Extended White-box robustness plots of ResNet-18 adversarially trained on CIFAR-10. Rows represent adversarial sample generation method used for adversarial training (Row-1) FGSM, (Row-2) FGSM-LL, and (Row-3) FGSM-Rand. Columns denote different adversary attacking methods such as, (Column-1) FGSM, (Column-2) FGSM-LL and (Column-3) FGSM-Rand.

## 9 Ensemble Adversarial Training ($EAT$)

In this section we show Extended Black-box robustness plots of networks trained using ensemble adversarial training for experiments presented in section 4.4 of the main paper. Table 4 shows the setup used for ensemble adversarial training. Intermediate models evolved during training of the held-out model are used to generate attacks. Figure 10 shows the Extended Black-box robustness plot of networks trained on CIFAR-100 and CIFAR-10 datasets using $EAT$. It is observed that models trained using $EAT$ are susceptible (e.g., $15-20\%$ drop in recognition accuracy for CIFAR-100) to the adversaries generated by intermediate models of the held-out model (observe towards the left part of the plot).

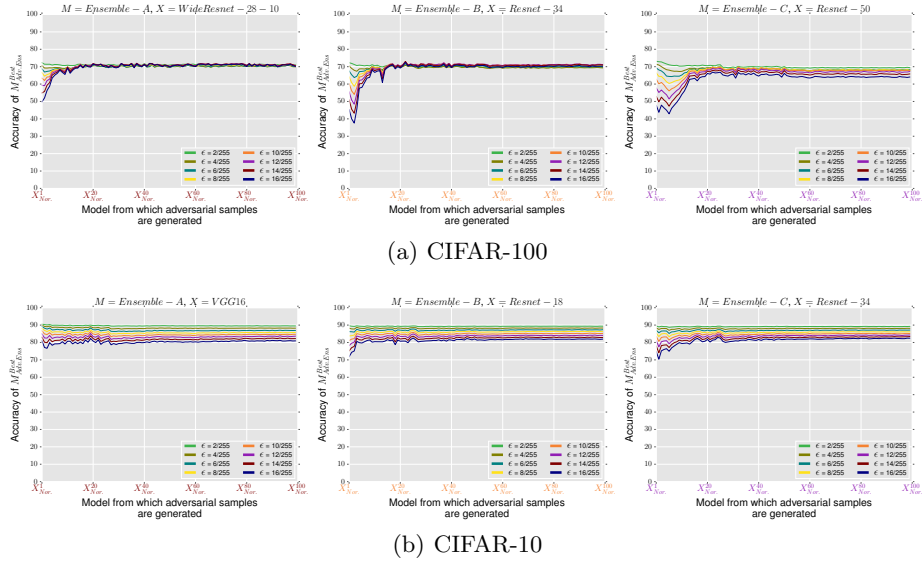

Fig. 10: Extended Black-box robustness plots of networks trained using *Ensemble Adversarial training (EAT)*. (a) Networks trained on CIFAR-100, (b) Networks trained on CIFAR-10.

Table 4: Setup used for Ensemble adversarial training (EAT). Refer to table 5 for MNIST network.

|  | **Network to be trained** | **Pre-trained Models** | **Held-out Model** |
|---|---|---|---|
| CIFAR-100 | WideResnet-28-10(Ensemble A) | Resnet-50, Resnet-34 | WideResnet-28-10 |
|  | WideResnet-28-10(Ensemble B) | WideResnet-28-10, Resnet-50 | Resnet-34 |
|  | WideResnet-28-10(Ensemble C) | WideResnet-28-10, Resnet-34 | Resnet-50 |
| CIFAR-10 | Resnet-34(Ensemble A) | Resnet-34, Resnet-18 | VGG-16 |
|  | Resnet-34(Ensemble B) | Resnet-34, VGG-16 | Resnet-18 |
|  | Resnet-34(Ensemble C) | Resnet-18, VGG-16 | Resnet-34 |
| MNIST | A(Ensemble A) | A,B,C | D |
|  | B(Ensemble B) | B, C ,D | A |
|  | C(Ensemble C) | C, D, A | B |
|  | D(Ensemble D) | D, A ,B | C |

## 10    Experiments on MNIST

In this section we show results on MNIST dataset, for experiments presented in the main paper. We use LeNet [11] for MNIST [10] dataset and for *Ensemble*

Table 5: Architecture of networks used in ensemble adversarial training on MNIST dataset.

| A | B | C | D |
|---|---|---|---|
| Conv(64,5,5) + Relu | Dropout(0.2) | Conv(128,3,3) + Tanh | $\left\{ \begin{array}{l} \text{FC(300) +Relu} \\ \text{Dropout(0.5)} \end{array} \right\} \times 4$ |
| Conv(64,5,5) + Relu | Conv(64,8,8) + Relu | MaxPool(2,2) | |
| Dropout(0.25) | Conv(128,6,6) + Relu | Conv(64,3,3) + Tanh | FC + Softmax |
| FC(128) + Relu | Conv(128,5,5) + Relu | MaxPool(2,2) | |
| Droupout(0.5) | Droupout(0.5) | FC(128) + Relu | |
| FC + Softmax | FC + Softmax | FC + Softmax | |

*Adversarial Training* we use networks described in Table 5 and ensemble setup is shown in Table 4. For training we used SGD with momentum, and models are saved for every 5 iterations. Images are pre-processed to be in $[0, 1]$ range and no data-augmentation is performed.

**NOTE:** Robustness plots of models trained on MNIST dataset are obtained by using intermediate models that are saved for every *5 iteration*. Since models trained on MNIST dataset achives high performance within one epoch.

### 10.1   Limitations of existing evaluation method

We adversarially train LeNet on MNIST dataset and while training, FGSM is used for adversarial sample generation process. After training, we obtain Extended White-box robustness plot using FGSM attack. Figure 11 shows the obtained Extended White-box robustness plot. It can be observed that adversarially trained models are not robust to the adversaries generated by the intermediate models, which the existing way of evaluation fails to capture. Observe dip in the accuracy of the $M_{adv}^{Best}$ on the left side of the plot.

### 10.2   Which attack to use for adversarial training, FGSM, FGSM-LL or FGSM-Rand

We perform normal training of LeNet on MNIST dataset and obtain its corresponding Extended White-box robustness plots using FGSM, FGSM-LL, and FGSM-Rand attacks respectively. Figure 12 shows the obtained plots. Note that the plots are towards the lower recognition accuracy for FGSM attack (column-1) compared to others. Hence, it is clear that FGSM attack (column-1) produces stronger attacks compared to FGSM-LL (column-2) and FGSM-Rand (column-3) attacks. In other words, drop in the model's accuracy is more for FGSM attack.

Additionally, we adversarially train the above network using FGSM, FGSM-LL and FGSM-Rand respectively for adversarial sample generation process. After training, we obtain robustness plots with FGSM, FGSM-LL and FGSM-Rand attacks respectively. Figure 13 shows the obtained Extended White-box robustness plots for all the three versions of adversarial training. It is observed that the model is more susceptible to FGSM attack (column-1) compared to other two attacks. Observe that for wide range of perturbations ($\epsilon$) FGSM attack (column-1) cause more dip in the accuracy.

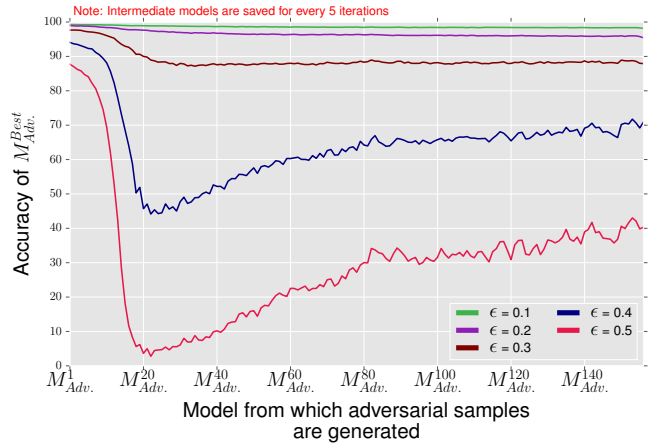

Fig. 11: Extended White-box robustness plot with FGSM attack, obtained for LeNet adversarially trained on MNIST dataset. Note that the classification accuracy of the best model ($M_{Adv}^{Best}$) is poor for the attacks generated by early models (towards origin) as opposed to that by the later models.

### 10.3    Gray-box adversarial training ($GAT$)

We train LeNet on MNIST dataset using the proposed Gray-box Adversarial Training (GAT) algorithm. In order to create strong adversaries, we chose FGSM to generate adversarial samples. We use the same set of hyper-parameters $D=$ 0.2, $EL=$ 0.5, and $T=$ ($1/4^{th}$ of an epoch). Specifically, we save intermediate models for every 0.2 drop ($D$) in the training loss till the loss falls below 0.5 ($EL$). Also, each of the saved models is sampled from the ensemble and generates adversarial samples for $1/4^{th}$ of an epoch. After training, we obtain Extended White-box robustness plots and Extended Black-box robustness plots for networks trained with the Gray-box adversarial training (Algorithm 1, main-paper) and for adversarially trained networks. Figure 14 shows the obtained robustness plots, it is observed from the plots that in the proposed gray-box adversarial training (row-2) there are no deep valleys in the robustness plots, whereas for the networks trained using existing adversarially training method (row-1) exhibits deep valley in the robustness plots. Table 6 presents the *worst-case accuracy* $A_w$ of the models trained using different training methods. Note that $A_w$ is significantly better for the proposed $GAT$ compared to the models trained with $AT$ and $EAT$ for a wide range of attack strength ($\epsilon$).

### 10.4    Ensemble adversarial training ($EAT$)

In this subsection, we compare our Gray-box Adversarial Training against Ensemble Adversarial Training [21] for networks trained on MNIST dataset. Ensemble adversarial training uses fixed set of pre-trained models for generating adversarial samples along with the current state of the network. For each iteration during training, the source model for generating adversaries is picked at random among the current and ensemble models. Table 4 shows the setup used for ensemble adversarial training, which contains network to be trained, pre-trained source models used for generating adversarial samples and the held-out model used for black-box attack. Figure 15 shows the Extended White-box robustness plots for the networks trained using ensemble adversarial training algorithm. Note the presence of deep and wide valleys in the plot. Whereas, the Extended White-box robustness plots for the models trained using the Gray-box adversarial training shown in figure 14 (row:2,column:3), do not have deep and wide valley. Figure 16 shows the Extended Black-box robustness plots obtained for networks trained using $EAT$ algorithm, ensemble A and D are more susceptible to Extended Black-box attack.

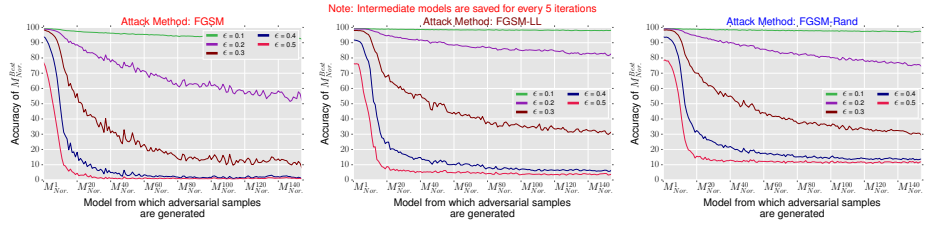

Fig. 12: Extended White-box robustness plots of LeNet normally trained on MNIST dataset, obtained using FGSM (column-1), FGSM-LL (column-2) and FGSM-Rand (column-3) attacks. Note that for a wide range of perturbations, FGSM attack causes more dip in the accuracy.

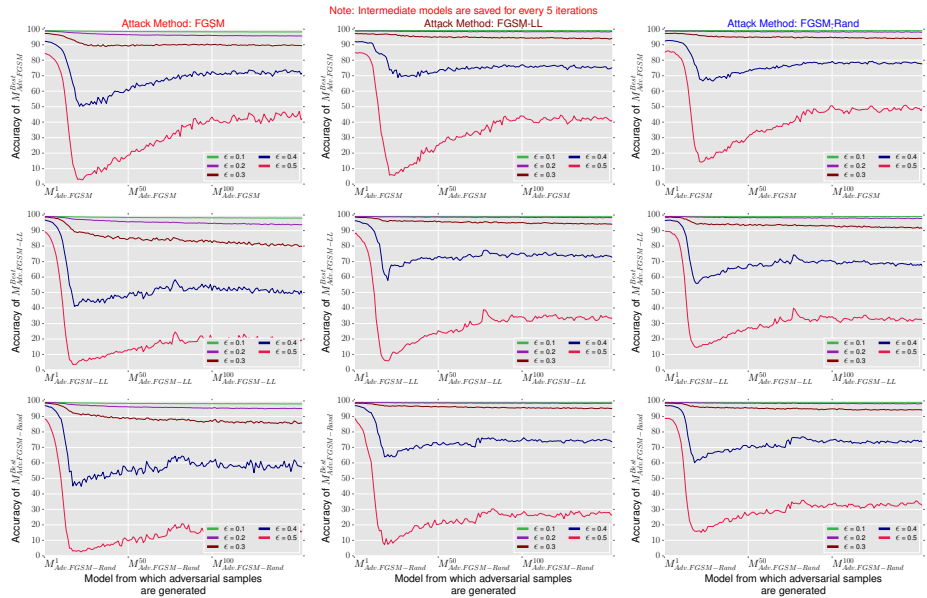

Fig. 13: Extended White-box robustness plots of LeNet adversarially trained on MNIST. Rows represent adversarial sample generation method used for adversarial training (Row-1) FGSM, (Row-2) FGSM-LL, and (Row-3) FGSM-Rand. Columns denote different adversary attacking methods such as, (Column-1) FGSM, (Column-2) FGSM-LL and (Column-3) FGSM-Rand.

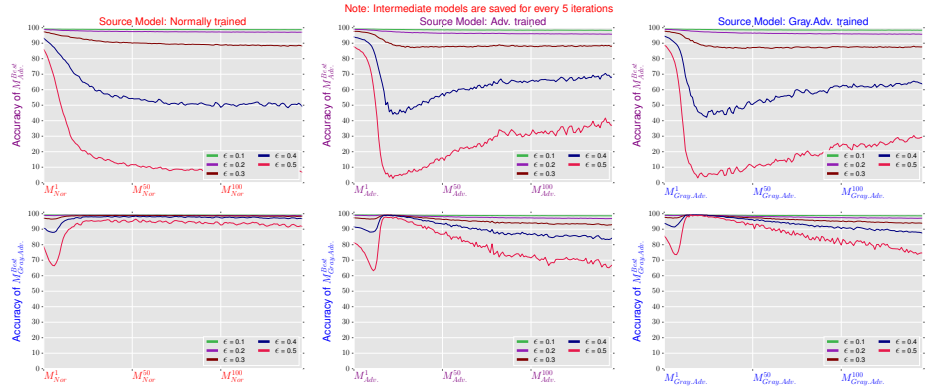

Fig. 14: Robustness plots of LeNet trained on MNIST dataset, obtained using FGSM attack. Rows represents the training method, Row-1:Model trained adversarially using FGSM and Row-2:Model trained using Gray-box adversarial training. Adversarial samples are generated by intermediate models of Column-1:Normal training, Column-2: Adversarial training using FGSM, Column-3:Gray-box adversarial training.

Table 6: Worst case accuracy $A_w$ of models trained on MNIST-10 using different training methods. For ensemble adversarial training ($EAT$) refer to section 10.4.

| Training Method | $A_w$ for various $\epsilon$ | | | | |
|---|---|---|---|---|---|
| | $\epsilon = 0.1$ | $\epsilon = 0.2$ | $\epsilon = 0.3$ | $\epsilon = 0.4$ | $\epsilon = 0.5$ |
| Normal | 93.21 | 55.99 | 11.89 | 2.16 | 0.89 |
| Adversarial | 98.91 | 97.42 | 88.95 | 44.15 | 2.77 |
| $EAT$ A | 99.09 | 97.22 | 88.69 | 60.73 | 23.45 |
| $EAT$ B | 98.37 | 95.24 | 74.03 | 42.19 | 38.45 |
| $EAT$ C | 98.62 | 95.45 | 81.7 | 50.06 | 19.99 |
| $EAT$ D | 95.07 | 87.17 | 61.03 | 26.12 | 10.65 |
| $GAT$(ours) | **99.25** | **98.74** | **96.6** | **88.31** | **63.17** |

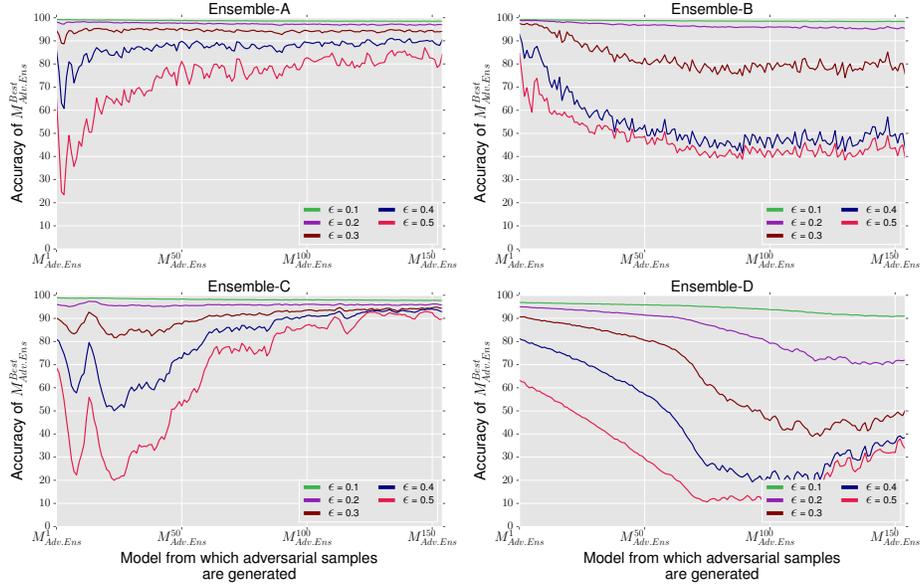

Fig. 15: Extended White-box robustness plots of models trained on MNIST dataset using ensemble adversarial training algorithm ($EAT$).

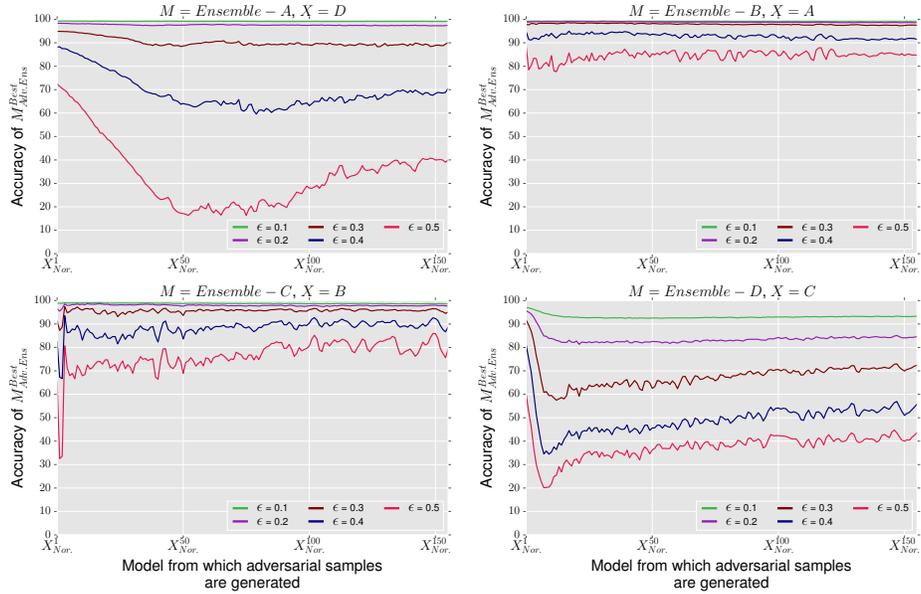

Fig. 16: Extended Black-box robustness plots of networks trained on MNIST dataset using *Ensemble Adversarial training (EAT)*.